\documentclass[10pt,twocolumn,letterpaper]{article}

\usepackage{iccv}
\usepackage{times}
\usepackage{epsfig}
\usepackage{graphicx}
\usepackage{amsmath}
\usepackage{amssymb}
\usepackage[accsupp]{axessibility}

\usepackage{bm}
\usepackage{bbm}
\usepackage{amsthm,amsmath,amssymb}
\usepackage{mathrsfs}
\usepackage{booktabs}
\usepackage{multirow}
\usepackage{bbding}
\usepackage{makecell}
\usepackage[switch]{lineno}  %
\usepackage{romannum}
\usepackage{enumitem}

\usepackage[breaklinks=true,bookmarks=false,colorlinks]{hyperref}

\iccvfinalcopy 

\def\iccvPaperID{4145} 

\ificcvfinal\fi

\begin{document}

\title{Spatio-Temporal Dynamic Inference Network for Group Activity Recognition}
\author{Hangjie Yuan\textsuperscript{\rm 1} \quad Dong Ni\textsuperscript{\rm 1,2}\thanks{Corresponding author.} \quad Mang Wang\textsuperscript{\rm 3} \\

\textsuperscript{\rm 1}College of Control Science and Engineering, Zhejiang University, Hangzhou, China \\ 
\textsuperscript{\rm 2} State Key Laboratory of Industrial Control Technology, Zhejiang University, Hangzhou, China \\
\textsuperscript{\rm 3}DAMO Academy, Alibaba Group, China\\
\tt\small \{hj.yuan,dni\}@zju.edu.cn, wangmang.wm@alibaba-inc.com\\
}





\maketitle
\ificcvfinal\thispagestyle{empty}\fi

\begin{abstract}
    Group activity recognition aims to understand the activity performed by a group of people. In order to solve it, modeling complex spatio-temporal interactions is the key. Previous methods are limited in reasoning on a predefined graph, which ignores the inherent person-specific interaction context. Moreover, they adopt inference schemes that are computationally expensive and easily result in the over-smoothing problem. In this paper, we manage to achieve spatio-temporal person-specific inferences by proposing Dynamic Inference Network (DIN), which composes of Dynamic Relation (DR) module and Dynamic Walk (DW) module. We firstly propose to initialize interaction fields on a primary spatio-temporal graph. Within each interaction field, we apply DR to predict the relation matrix and DW to predict the dynamic walk offsets in a joint-processing manner, thus forming a person-specific interaction graph. By updating features on the specific graph, a person can possess a global-level interaction field with a local initialization. Experiments indicate both modules' effectiveness. Moreover, DIN\footnote{Codes are available at \url{https://github.com/JacobYuan7/DIN_GAR}.} achieves significant improvement compared to previous state-of-the-art methods on two popular datasets under the same setting, while costing much less computation overhead of the reasoning module. 
\end{abstract}

\section{Introduction}
Group activity recognition (GAR) aims to infer an overall activity performed by a group of people in the scene \cite{choi2009they,ibrahim2016hierarchical,bagautdinov2017social,yan2018pctdm,qi2018stagnet,wu2019learning,gavrilyuk2020actor,yuan2021learningcontext}. It has aroused research interests due to various applications, including surveillance/sports video analysis, social scene understanding, \textit{etc.} The critical problem that lies in GAR is to infer a group-level activity representation given a video clip, which asks for elaborately designed reasoning modules.

Recently proposed reasoning modules mainly incorporate spatio-temporal interactive factors to get a refined activity representation. Modeling of agents' interactions has been widely studied. The mostly adopted methods are recurrent neural networks \cite{alahi2016sociallstm,zellers2018neuralmotif}, the attention mechanism \cite{vemula2018socialattention,hoshen2017vain} and graph neural networks (GNNs) \cite{sun2019stochasticprediction,fu2019dual,xu2017iterativemessage}. GNNs have been a frequently adopted method in GAR \cite{qi2018stagnet,wu2019learning,yan2020higcin,pramono2020eccvempowering}, which performs message passing on a constructed semantic graph and achieves competitive results on publicly available benchmarks.



\begin{figure}[t]
\centering
\includegraphics[width=0.47\textwidth]{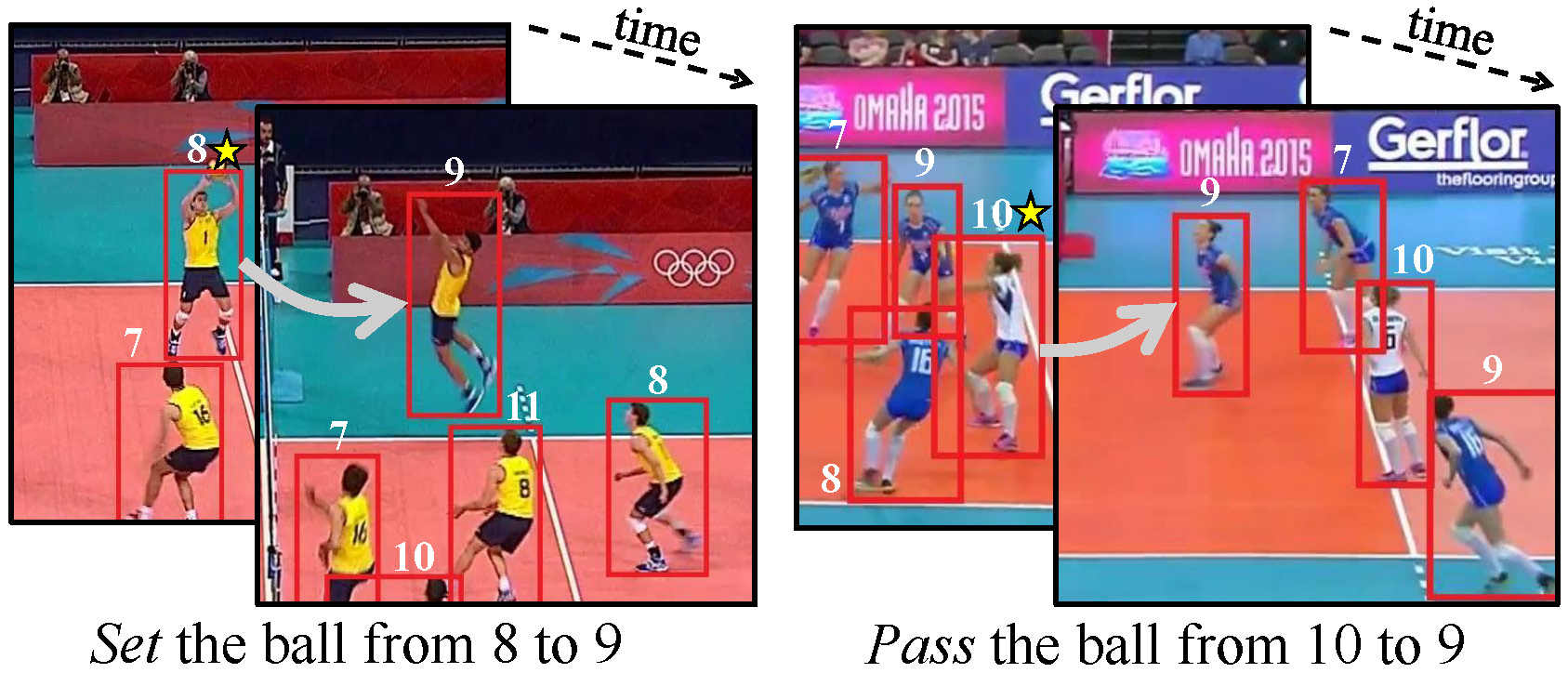} 
\caption{\textbf{Examples of \textit{right set} and \textit{right pass} group activity}. The red bounding box annotated with a star is the person performing the key action for the activity. The grey arrow denotes the key interaction linking the starred person and the semantically important person, which is always not aligned in the spatial or temporal domain. The person indices do not start from 1 because we only illustrate part of the images.}
\label{pass_set_example}
\vspace{-0.4cm}
\end{figure}

\begin{figure}[t]
\centering
\includegraphics[width=0.47\textwidth]{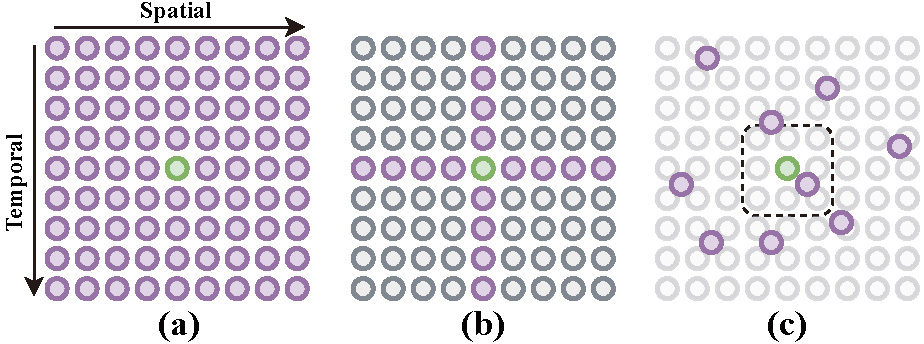} 
\caption{\textbf{Visualizations of three inference schemes} in spatio-temporal domain with GNNs for GAR. The green node denotes the feature to be updated. The purple nodes denote features involved in updating the green node. (a) Fully-connected graph inference. (b) Criss-cross graph inference. (c) Proposed person-specific dynamic graph inference, which is unique for every green node. The dashed box is an example of an initialized interaction field.}
\label{inference_scheme}
\vspace{-0.4cm}
\end{figure}

However, previous methods using GNNs stick to a paradigm that models the interaction between individuals on a predefined graph as shown in Figure \ref{inference_scheme}. It is a feasible way but bears several drawbacks: \textbf{i)} Those who interact with a given person should be person-specific but not predefined. Like in Figure \ref{pass_set_example}, a person will interact with people depending on their own context: the 8th person in the left video interacts with the 9th person who is about to spike the ball; the 10th person in the right video interacts with the 9th person who is about to set the ball. A predefined graph can not suit every person's inference. \textbf{ii)} Previous predefined graph models infer interactions on a fully-connected \cite{wu2019learning} or criss-cross \cite{yan2020higcin,pramono2020eccvempowering} graph which is shown in Figure \ref{inference_scheme}(a) and (b). It easily results in the over-smoothing \cite{deeper_insignt_intoGCN} that makes features indistinguishable and damages the performance. Also, it costs overmuch computation overhead if expanding to long video clips or expanding to a scenario with too many people in the scene.

Aiming at solving the drawbacks mentioned above, inspired by \cite{dai2017dcnv1,zhang2020DGMN},we present Dynamic Inference Network (DIN), which contains Dynamic Relation (DR) and Dynamic Walk (DW). These two modules combined can predict a person-specific interaction graph for better modeling interactions as shown in Figure \ref{inference_scheme}(c). For a given person feature on a spatio-temporal graph, we set a spatio-temporal interaction field around it as an initialization, which is shared between DR and DW. This interaction field determines the people to be involved in inferring the interaction graph. The initialized field size will not increase if the spatial or temporal axis expands, which reduces the computation.

Within this initialized interaction field, we use DR to predict a person-specific relation matrix, denoting the interaction relations between persons. The features in the interaction field endow the relations with an interaction context. Then, to facilitate the model to learn from complex spatio-temporal interactions, we use DW to predict dynamic walk offsets for every feature within the field. The dynamic walks allow for the locally initialized interaction field to form a graph that enables global-level interactions. The proposed modules are easy for deployment onto any widely used backbones to form a pipeline named DIN. Besides, previous methods seldom make computational complexity analysis, which is a significant evaluation for a designed module. In this paper, we present computational complexity analysis and show that our modules cost less computation overhead while performing better.


To summarize, our contributions are listed as follows: 
\begin{itemize}[noitemsep,topsep=0pt,parsep=0pt,partopsep=0pt]
    \item We propose DIN to construct person-specific interaction graphs in the spatio-temporal domain, which are not predefined and can also serve as a general approach for modeling interactions.  
    \item We propose DR that predicts person-specific relation matrices and DW that allows for the locally initialized interaction field to update features globally. Both are proved useful by experiments.
    \item We prove by experiments that a small size of initialized interaction field is sufficient for existing datasets. We use a case visualization to exemplify that interaction graphs can capture the key person and key interactions, and a locally initialized interaction field can cover a global-level interaction field with proposed modules. 
    \item DIN achieves state-of-the-art performances under the setting of the same backbone and input modality on two widely used benchmarks, while costing much less computation overhead of the reasoning module. 
\end{itemize}

%



\section{Related Work}
\begin{figure*}[t]
\centering
\includegraphics[width=1\textwidth]{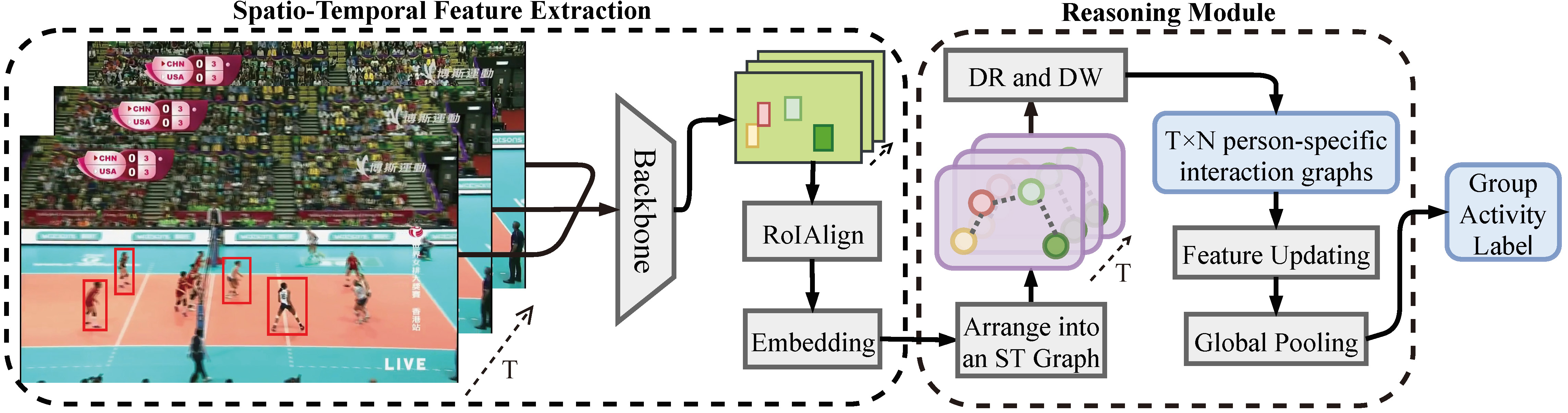} 
\caption{\textbf{The overall pipeline of Dynamic Inference Network}. Generally, it consists of two stages: i) Spatio-temporal feature extraction, ii) Reasoning module. Note that there will be $T \times N$ unique interaction graphs for updating. In our codebase, the first stage is shared with previous methods. The main variations are in the Reasoning Module. We only illustrate 4 bounding boxes in the image for clarity.}
\label{overall_pipeline}
\vspace{-0.4cm}
\end{figure*}

\textbf{Group Activity Recognition} \ Group activity recognition was firstly proposed in \cite{choi2009they}. Following works \cite{choi2011learningcontext,lan2011discriminative,lan2012social,choi2012aunifiedframework,amer2014hirf} were basically to extract hand-crafted features (\textit{e.g.}, HOG \cite{dalal2005HOG}) and apply graphical models to infer group activity representations. With the boom of deep learning, methods incorporating convolution neural networks (CNNs) and recurrent neural networks (RNNs) have proved effective. For example, the works of \cite{ibrahim2016hierarchical,bagautdinov2017social} managed to model the temporal dynamics in action level or group level via RNNs on CNN features. The works of \cite{wang2017merge_two_class,yan2018pctdm,qi2018stagnet,2019hierarchical_longshortterm} applied RNNs to the modeling of person interactions. The attention mechanism also proved its effectiveness in GAR. The works of \cite{yan2018pctdm,qi2018stagnet,tang2019coherence} combined RNNs with attention mechanisms to capture the key features in the spatial or temporal domain. Specifically, the self-attention mechanism was introduced to learn the temporal evolution and spatial interactions \cite{gavrilyuk2020actor,pramono2020eccvempowering}. 


GNNs, which inferred on graph-structured data, attracted researchers' attention in GAR. ARG \cite{wu2019learning} firstly proposed to use graph convolution networks (GCNs) to learn person interactions on a spatio-temporal graph. Later, several works \cite{yan2020higcin,pramono2020eccvempowering,weaklyGAR} improved the previous fully-connected graph to a criss-cross one when modeling relations and aggregating features. However, they all ignored the person-specific interaction context. Our work is partly inspired by deformable convolution \cite{dai2017dcnv1,zhu2019dcnv2}, whose relations are not conditioning on the person features. Moreover, related work like DGMN \cite{zhang2020DGMN} which mentioned 'dynamic' constrained in implicit pixel-level spatial feature enhancement, while our pipeline suits video processing and agent-level spatio-temporal reasoning.




\textbf{Modeling of Interactions} \ The modeling of interactions is significant in understanding a complex system with multiple objects/agents \cite{battaglia2016interactionnetwork,chang2016learningphysicaldynamics}. Many research areas inherently involve the modeling of interactions like trajectory prediction \cite{gupta2018socialgan,sun2019stochasticprediction,sadeghian2019sophie}, human object interaction \cite{qi2018hoibygpnn,li2019transferable,gao2020drg} and scene graph generation \cite{xu2017iterativemessage,zellers2018neuralmotif,chen2019kern}. In GAR, modeling interactions is crucial in understanding their overall activity \cite{bagautdinov2017social}. Among their adopted methods, GNNs have been a frequently chosen method. Some related works like EvolveGCN \cite{pareja2020evolvegcn} explored a better representation learning strategy on evolving graphs and EvolveGraph \cite{li2020evolvegraph} explored a prediction method to adjust the structure of one graph. However, we focus on the exploration of constructing dynamic agent-specific graphs based on their interaction field. The proposed modules are general approaches to tackle the modeling of interactions in related problems.

\section{Method}
In this section, we firstly outline the pipeline of DIN. Then, we give a brief review of previous GNN reasoning modules for GAR. Finally we introduce the modules that we propose to dynamically infer the group activity. To better present the idea, we specifically present the feature updating method for the $i$th person feature.

\subsection{Dynamic Inference Network}
The integrated framework, which we denote as Dynamic Inference Network (DIN), is illustrated in Figure \ref{overall_pipeline}. The DIN takes in a short clip of videos, which is fed into a selected backbone network to extract visual features. For the backbone network, we mainly experiment on ResNet-18 \cite{he2016resnet} and VGG-16 \cite{simonyan2014vgg} to demonstrate the effectiveness of our proposed module and to seek for a fair comparison with previous methods. Then RoIAlign \cite{he2017mask} is applied to extract the person features aligned with bounding boxes, which are then embedded to a $D$-dimension space. We stack the person features to form $\bm{X} \in \mathbb{R}^{T \times N \times D}$, where $T,N$ denotes the temporal steps (\textit{i.e.}, temporal dimension) and number of annotated people in each frame (\textit{i.e.}, spatial dimension) respectively. Note that the spatial dimension is ordered by people's coordinates following \cite{wu2019learning,gavrilyuk2020actor}. It is then arranged into a Spatio-Temporal graph (ST graph). The proposed DR and DW dynamically predict a specific interaction graph for a selected feature ($T \times N$ interaction graphs in total). Thus, we can operate feature updating accordingly.

After the above inference, we can perform a global pooling to get the final group representation, which contains a max-pooling layer along the spatial dimension and an average pooling layer along the temporal dimension. The training objective is the cross-entropy loss for group activities. Although many previous methods like \cite{qi2018stagnet,wu2019learning,azar2019crm,gavrilyuk2020actor} use additional cross-entropy loss for individual actions, the action labels are actually ill-defined \cite{weaklyGAR} and expensive in labeling. We use cheap group activity labels while still achieving competitive results. 

Although computational complexity analysis in Section \ref{experiments} already indicates that DR and DW bring in limited parameters and FLOPs apart from the backbone and the embedding layer, we take a step further to seek for a lighter reasoning module. In practice, we apply pointwise convolution \cite{howard2017mobilenetsv1} before the reasoning module to reduce the dimension of $\bm{X}$ from $D$ to $D_l$. We name this model \textbf{Lite DIN}.


\subsection{Recap of Previous GNN Reasoning} \label{recap}

\begin{figure*}[t]
\centering
\includegraphics[width=1\textwidth]{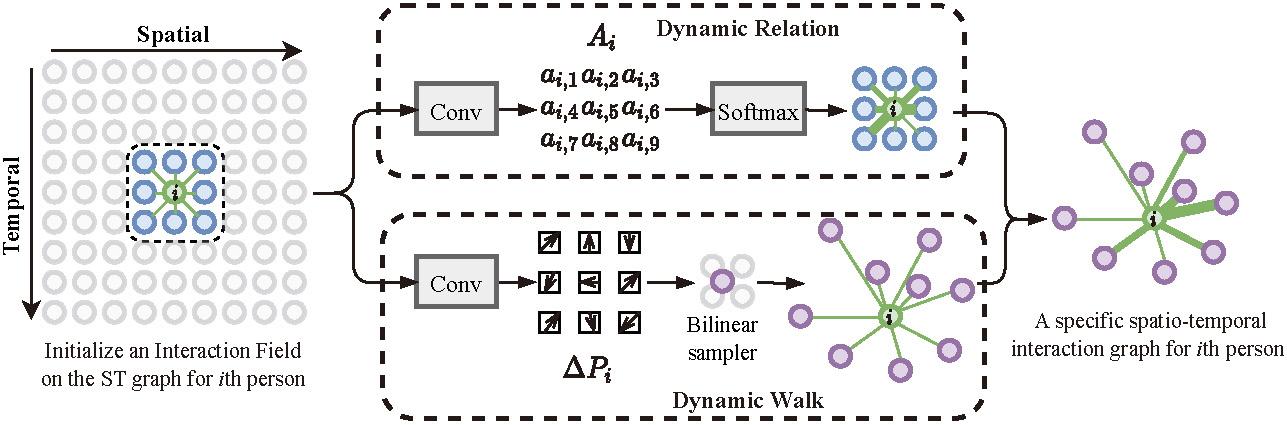} 
\caption{\textbf{Details of DR and DW} on creating the person-specific interaction graph for $i$th person. For the given person, DR predicts a relation matrix and DW predicts the dynamic walk offsets to endow the interaction graph with a global interaction field, both based on an initialized interaction field (we set it to $3\times3$ as an example).}
\label{DR_DW_details}
\vspace{-0.4cm}
\end{figure*}


We start the recap by introducing Actor Relation Graph (ARG) in \cite{wu2019learning}. The spatio-temporal feature extraction stage for ARG is identical to the DIN as illustrated in Figure \ref{overall_pipeline}. It uses a fully-connected graph as illustrated in Figure \ref{inference_scheme}(a). The spatial and temporal dimension of person features $X$ are collapsed to one, denoted as $\bm{X} = \{\bm{x}_i\}_{i=1}^{TN}$ with $ \bm{x}_i \in \mathbb{R}^{D}$. Their pairwise relations can be denoted as $\bm{R} = \{r_{i,j}| i,j = 1,...,TN\}$ with $r_{i,j} \in \mathbb{R}^{1}$, which can be computed by
\begin{equation} \label{embed_dot_product}
r_{i,j} = \frac{\theta(\bm{x}_i)^{\rm T} \phi(\bm{x}_j)}{\sqrt{D_r}}  
\end{equation}
\vspace{-0.2cm}
\begin{equation}
\Tilde{r}_{i,j} = {\rm softmax}_j(r_{i,j}) \
= \frac{{\rm exp}(r_{i,j})}{\sum_{q=1}^{TN}{{\rm exp}(r_{i,q})}}   \label{softmax}
\end{equation}
where $\theta$ and $\phi$ are linear transformations functions, \textit{i.e.}, $\theta(\bm{x}_i) = \bm{W}_{\theta}\bm{x}_i$ with $\bm{W}_{\theta} \in \mathbb{R}^{D_r \times D}$ and $\phi(\bm{x}_i)$ is similarly defined; $D_r$ is the dimension of the embedding space; ${\rm softmax}_j$ defines a softmax function along the index $j$ to get the normalized relations $\Tilde{r}_{i,j}$. We do not formulate distance mask here for clarity.

We perform one-layer ARG to update the person feature as 
\begin{equation}
\bm{x}_{i}^{(l+1)} = \sum\nolimits_{g=1}^{N_g}\sigma \left(  
    \sum\nolimits_{j=1}^{TN}{\Tilde{r}_{i,j} \bm{x}_j^{(l)} \bm{w}^{(g)}} \right) + \bm{x}_i^{(l)}
\end{equation}
where $N_g, g, l$ denote the number of graphs in one layer, the graph index and the layer index respectively; $\sigma$ is an activation function (${\rm ReLU}$ in our implementation); $\bm{w}^{(g)} \in \mathbb{R}^{D \times D}$ is the graph-specific trainable transformation matrix. Note that $\bm{w}^{(g)}$ and $\Tilde{r}_{i,j}$ are also layer-specific but for the purpose of clarity, we omit this superscript $l$. Similar thing is done for learnable parameters and relations in following equations.

After the feature updating, we finally perform a global pooling operation on the reshaped $\bm{X}^{(l+1)} \in \mathbb{R}^{T \times N \times D}$ to get the final group representation $\bm{z} \in \mathbb{R}^{D}$. The Cross Inference Block proposed in \cite{yan2020higcin} ameliorates the fully-connected inference by criss-cross inference as shown in Figure \ref{inference_scheme}(b).


\subsection{Dynamic Relation} \label{DR}

Before we dive into the proposed modules, we present the definition for the interaction field. The \textbf{Interaction Field} is a region upon an ST graph that is involved in inferring the interaction features. One example of the interaction field is shown using a dashed box on the ST graph in Figure \ref{DR_DW_details}. Our proposed modules, \textit{i.e.} DR and DW, jointly process features within this field to infer person-specific interaction graphs. The initialized interaction field covers a selected person's spatio-temporal neighborhoods, which provides direct interaction cues. More complex initializations are left for future exploration.

We propose Dynamic Relation (DR) to infer the relation matrix for the person-specific interaction graph. An illustration of DR is shown in the upper branch of Figure \ref{DR_DW_details}. 'Dynamic' in DR refers to the fact that the relation matrix is dependent on the features in the initialized interaction field, rather than sticks to the same when updating every feature.

To infer the dynamic relations within this field, we adopt convolution following \cite{dai2017dcnv1,zhang2020DGMN}. For a selected $i$th feature on the original ST graph, we denote $\bm{u}_i \in \mathbb{R}^{(K \times D)}$ as the stacking features within its interaction field and denote $K$ as the interaction field size, \textit{e.g.}, $K=9$ if the interaction field is $3\times3$. We rewrite the convolution in a matrix form as
\begin{equation}
    \bm{A}_{i} = \bm{W}_a \bm{u}_i + \bm{b}_a
\end{equation}
where $\bm{W}_a \in \mathbb{R}^{K \times (K \times D)}$ is the linear projection matrix for inferring relations; $\bm{b}_a \in \mathbb{R}^{K}$ is the bias parameters. $\bm{A}_i = \{a_{i,k}|k=1,...,K\}$ with $a_{i,k} \in \mathbb{R}^{1}$ is the relation matrix for $i$th feature, where $k$ enumerates $K$ features in $i$th feature's initialized field. Similar to Eq.\ref{softmax}, $\Tilde{a}_{i,k}$ is the normalized $a_{i,k}$ along the index $k$, \textit{i.e.}, $\Tilde{a}_{i,k} = {\rm softmax}_k(a_{i,k})$.


Instead of updating the features in a fully-connected graph or criss-cross graph, we update the features within the initialized field as
\begin{equation}
    \bm{x}_i^{(l+1)} = \sigma\left(
        \sum\nolimits_{k=1}^{K}{
        \Tilde{a}_{i,k} \bm{x}_k^{(l)} \bm{w}}\right) + \bm{x}_i^{(l)}
\end{equation}
Note that we do not incorporate multiple graphs due to its excessive parameters and trivial improvement \cite{wu2019learning}. 



\subsection{Dynamic Walk} \label{DW}

Although DR has successfully inferred their relations with all person features in the initialized interaction field, it still follows a predefined message passing route, which lacks the ability for person-specific interaction modeling. Moreover, previous methods manage to model long-range spatio-temporal dependency by a fixed graph in a fully-connected or criss-cross scheme, which consumes excessive computational resources. We propose a Dynamic Walk (DW) module that enables features within the interaction field to execute dynamic walks on the primary ST graph. An illustration of DW is shown in the lower branch of Figure \ref{DR_DW_details}. Through DW, we hope to model complex spatio-temporal dependency using a size-limited interaction field. 'Dynamic' in DW refers to the fact that the interaction graph is dependent on the features in the initialized interaction field, which is not predefined anymore. 

To allow for dynamic walk, we need to predict their spatio-temporal dynamic walk offsets. For a selected $i$th person feature, we denote the dynamic walk offsets for all features within the interaction field as $\Delta \bm{P}_i = \{\Delta \bm{p}_{i,k}|k=1,...,K\}$ with $\Delta \bm{p}_{i,k} \in \mathbb{R}^{2}$. We predict the dynamic walk offsets as 
\begin{equation}
    \Delta \bm{P}_i = \bm{W}_p \bm{u}_i + \bm{b}_p
\end{equation}
where $\bm{W}_p \in \mathbb{R}^{(K \times 2) \times (K \times D)}$ is the linear projection matrix for predicting dynamic walk offsets; $\bm{b}_p \in \mathbb{R}^{(K \times 2)}$ is the bias term. Similar to DR, this predicts the dynamic walk offsets for all features within the field and it is instantiated by convolution. Using the predicted offsets, we can obtain the dynamic-walked features by performing dynamic walk on the ST graph. Note that the dynamic-walked features are clamped to be within the range of the ST graph. As the dynamic walk offsets are constantly fractional, a bilinear sampler \cite{jaderberg2015spatialtransformer} is adopted to sample dynamic-walked features. We denote the coordinate of the $k$th feature in the $i$th interaction field as $\bm{p}_{i,k} \in \mathbb{R}^{2}$. Dynamic-walked features $\bm{Y}_i = \{\bm{y}_{i,k}|k=1,...,K \}$ with $\bm{y}_{i,k} \in \mathbb{R}^{D}$ can be formulated as
\begin{equation}
\begin{split}
    \bm{y}_{i,k} = \sum_{m=1}^{T}{
        \sum_{n=1}^{N}{\bm{x}_{(m-1)N+n} \delta(m, n, \bm{p}_{i,k}, \Delta \bm{p}_{i,k})}} 
\end{split}
\end{equation}
\vspace{-0.5cm}
\begin{equation}
\small
\begin{split}
    \delta(m, n, \bm{p}_{i,k}, \Delta \bm{p}_{i,k}) = {\rm max}(0, 1-|m-\bm{p}_{i,k}^T-\Delta \bm{p}_{i,k}^T|) \\ \times {\rm max}(0,1-|n- \bm{p}_{i,k}^N-\Delta \bm{p}_{i,k}^N|)
\end{split}
\end{equation}
where superscript '$T$' and '$N$' denotes the temporal and spatial coordinate. 

Based on the dynamic-walked features, we can update the $i$th feature as
\begin{equation}
    \bm{x}_i^{(l+1)} = \sigma\left(
        \sum\nolimits_{k=1}^{K}{
            \Tilde{a}_{i,k} \bm{y}_{i,k}^{(l)} \bm{w}}\right) + \bm{x}_i^{(l)} 
\end{equation}
Note that in the above formulation, we combine the DR and DW to form the final dynamic updating function. 




\section{Experiments} \label{experiments}
In this section, we first present datasets and implementation details for GAR. Next, we perform quantitative analysis to explore the contributions of our modules and variances of different interaction field initializations, and to prove the superiority in terms of computational complexity. Then, we compare our methods with previous state-of-the-art methods. Finally, we provide visualizations to understand DIN better.

\subsection{Experiment Settings}
\textbf{Datasets} \ So far, there are two widely used datasets in group activity recognition, namely Volleyball dataset (VD) \cite{ibrahim2016hierarchical} and Collective Activity dataset (CAD) \cite{choi2009they}. 

The Volleyball dataset comprises of 3,493 training clips and 1,337 testing clips, which are trimmed from 55 videos of volleyball matches. For each short clip, it provides three kinds of annotations: i) coordinates of players' bounding boxes in the center frame of a given clip; ii) individual action labels for the annotated person: \textit{blocking, digging, falling, jumping, moving, setting, spiking, standing} and \textit{waiting}, which are not used in our experiments; iii) group activity labels for the given clip: \textit{right set, right spike, right pass, right winpoint, left set, left spike, left pass} and \textit{left winpoint}. To perform feature extraction on the whole clip, we use the tracklets provided by \cite{bagautdinov2017social}. Two metrics are used for evaluating the performance of a model, \textit{i.e.}, MCA (\%) which is short for Multi-class Classification Accuracy and MPCA (\%) which is short for Mean Per Class Accuracy.

The Collective Activity dataset comprises of 44 videos containing varying number of frames from 194 to 1,814 frames. Similar to VD, it is labelled with three levels of annotations: i) coordinates of people's bounding boxes on the center frame of every ten frames; ii) individual action labels for the annotated person: \textit{NA, crossing, waiting, queueing, walking} and \textit{talking}, which are not used in our experiments; iii) group activity labels for every ten frames: \textit{ crossing, waiting, queueing, walking} and \textit{talking}. We follow \cite{wang2017merge_two_class,yan2018pctdm,yan2020higcin} to merge the class \textit{crossing} and \textit{walking} into \textit{moving}. Similarly, we use the tracklets from \cite{bagautdinov2017social}. Train-test split follows \cite{qi2018stagnet}. MPCA is used for evaluation on this dataset due to class imbalance.

\textbf{Implementation Details} \  For VD, we use video images with resolution $H\times W = 720\times 1280$. For CAD, we use video images with resolution $H\times W = 480\times 720$. For both datasets, we use video clips which contain $T=10$ frames each following \cite{wu2019learning,yan2020higcin,qi2018stagnet,hu2020prl}. The maximum number of people in the scene is $N=12$ for VD and $N=13$ for CAD. We use person feature with embedding dimension $D = 1024$. For Lite DIN, we use the embedding dimension $D_l = 128$. The convolution operations for DR and DW are initialized by zero vectors \cite{dai2017dcnv1}. When applying convolution on the graph, we use zero paddings to maintain a fixed interaction field size. We follow \cite{wu2019learning} to initialize the backbone of DIN model with parameters from the base model. We do not use any action label supervision. For the training of VD, we employ Adam optimizer whose learning rate starts with $1\times 10^{-4}$ and decay rate is $\frac{1}{3}$ every 10 epochs. For the training of CAD, we employ the same optimizer whose learning rate starts and stays with $5\times 10^{-5}$. We run 30 epochs in total. The hyper-parameter for Adam is $\beta_1 = 0.9, \beta_2 = 0.999$ and $\epsilon = 10^{-8}$. 

\subsection{Quantitative Analysis} \label{quantitative_analysis} 
In this subsection, we conduct experiments on VD. We set the backbone for quantitative analysis to ResNet-18.


\begin{table}[t]
\small
  \centering
    \begin{tabular}{c|cc}
    \Xhline{1.0pt}
    Model & MCA & MPCA \\
    \hline\hline
    Base model & 87.8  & 88.4  \\
    DIN w/ DR & 92.1  & 92.3  \\
    DIN w/ DW & 92.0  & 92.5  \\
    DIN w/ DR+DW & \textbf{93.1}  & \textbf{93.3}  \\
    DIN w/ DR+DW\textsuperscript{*} & 92.9  & 93.1  \\
    \Xhline{1.0pt}
    \end{tabular}%
  \vspace{0.2cm}
  \caption{Ablation study on the usage of \textbf{DR and DW}. Experiments are conducted on VD. The backbone is set to ResNet-18.}
  \label{ablation_DR_DW}%
\end{table}%

\textbf{DR and DW} \ We first conduct ablation study to demonstrate the efficacy of proposed modules. We use a fixed initialized interaction field of $3 \times 3$ and following models: 
\begin{itemize}[noitemsep,topsep=0pt,parsep=0pt,partopsep=0pt]
    \item \textbf{Base model}: It consists of a backbone network, RoIAlign, the global pooling layer and a final classification layer.
    \item \textbf{DIN w/ DR}: It contains a backbone, RoIAlign, DR module, a global pooling layer and a classification layer. It allows for a relation matrix prediction within the interaction field.
    \item \textbf{DIN w/ DW}: It contains a backbone, RoIAlign, DW module, a global pooling layer and a classification layer. It allows for a dynamic walk prediction to expand its interaction field.
    \item \textbf{DIN w/ DR+DW}: It is defined analogously with above models. It allows for a dynamic relation prediction based on the original features in the field as illustrated in Figure \ref{DR_DW_details}.
    \item \textbf{DIN w/ DR+DW\textsuperscript{*}}: It is defined analogously with DIN w/ DR+DW, except that it allows for a dynamic relation prediction based on dynamic-walked features, \textit{i.e.}, $\bm{Y}_i$.
    
\end{itemize}
The results for the above models are shown in Table \ref{ablation_DR_DW}. The table indicates that incorporating any proposed modules can significantly improve the performance. Compared to ARG \cite{wu2019learning}, the result for DIN w/ DR indicates the superiority of joint processing and a small interaction field. The result for DIN w/ DW indicates the superiority of person-specific interaction graphs. We find that DIN w/ DR and DIN w/ DW show similar improvements compared to the base model. We consider it is because that after we perform the dynamic walk on a graph, the bilinear sampler defines the feature interpolation by bilinear weights, which is, to some extent, one kind of dynamic relations. These weights are determined by the dynamic walk offsets, which are not as straightforward as DR. The partial ability of dynamic relations and a global interaction graph enable DIN w/ DW to perform similarly with DIN w/ DR.

Combining DR and DW, the DIN model is endowed with more dynamicity and larger interaction fields, thus performing even better. Specifically, DIN w/ DR+DW performs slightly better than DIN w/ DR+DW\textsuperscript{*}, which indicates the initialized interaction field provides sufficient information for predicting relations. 

\textbf{Computational Complexity Analysis} \ In this subsection, we present the parameters and FLOPs that the reasoning module contains. Note that the reasoning module that we define does not include the backbone and the person feature embedding layer, as we mainly focus on an efficient reasoning module in this paper. Since previous methods' reported results vary in the input modality, backbones and implementation details, we re-implement them to fit into our framework and codebase while ensuring consistency to the original paper and their publicly available codes. We use an initialized interaction field of $3 \times 3$ for all our proposed modules. For a fair comparison, we set all their backbones to ResNet-18. The results are listed in Table \ref{comptational_complx}. Besides, we present the statistics for the backbone and person feature embedding as a reference: 24.8M \#Params, 674.6 GFLOPs for $720\times1280$ resolution and 24.8M \#Params, 254.9 GFLOPs for $480\times720$ resolution. The result has shown that the model with DR or DW alone achieves higher performances than previous methods, at the same time, reduces the computational complexity of the reasoning module. By combining DR and DW, our model benefits from the dynamicity of person-specific relation matrices and the dynamic walk offsets, thus achieving better performance. Note that the model with DR+DW adds very little computation cost compared to the model with DR or DW alone. The lite model with DR+DW achieves impressive results while adding very little computational overhead compared to the base model. 




\begin{table}[t]
\small
  \centering
    \begin{tabular}{c|cccc}
    \Xhline{1.0pt}
    {Reasoning Module} & \#Params & FLOPs & MCA & MPCA \\
    \hline\hline
    PCTDM \cite{yan2018pctdm} & 26.235M  & 6.298G  & 90.3  & 90.5  \\
    ARG \cite{wu2019learning}   & 25.182M  & 5.436G  & 91.1  & 91.4  \\
    AT \cite{gavrilyuk2020actor} & 5.245M  & 1.260G  & 90.0  & 90.2  \\
    HiGCIN\cite{yan2020higcin} & 1.051M  & 184.992G  & 91.4  & 92.0  \\
    SACRF \cite{pramono2020eccvempowering} & 29.422M  & 76.757G  & 90.7  & 91.0  \\
    \hline
    EDP    & 3.146M  & 0.755G  & 91.6  & 91.6  \\
    DR    & 1.140M  & 0.272G  & 92.1  & 92.3  \\
    DW   & 1.222M  & 0.291G  & 92.0  & 92.5  \\
    DR+DW & 1.305M  & 0.311G  & \textbf{93.1} & \textbf{93.3} \\
    Lite DR+DW & 0.180M  & 0.042G  & \textbf{92.6}  & \textbf{92.8}  \\
    \Xhline{1.0pt}
    \end{tabular}%
  \vspace{0.2cm}
  \caption{\textbf{Computational complexity analysis}. Their backbones are set to ResNet-18. \#Params and FLOPs for the backbone and the embedding layer are \textbf{not} included.}
  \label{comptational_complx}%
\end{table}%

Besides the proposed model variants, we further provide another reasoning module:
\begin{itemize}[noitemsep,topsep=0pt,parsep=0pt,partopsep=0pt]
    \item \textbf{EDP}: The corresponding model is DIN w/ EDP. It is similar to DIN w/ DR, except that it uses the Embedded Dot-Product (EDP, formulated in Eq.\ref{embed_dot_product}) within the interaction field rather than DR for inferring the relation matrix $\bm{A}_i$. We set $D_{r} = D$.
\end{itemize}
We can observe that \textbf{i)} the previous pairwise interaction model EDP which predicts relations using only two persons, performs slightly worse than DR and cost higher computation overhead; \textbf{ii)} If comparing EDP to ARG, it shows a small initialized field ameliorates the over-smoothing that ARG has due to fully-connected inference, and achieves better performances.






\begin{table}[t]
\scriptsize
\setlength{\tabcolsep}{1pt}
  \centering
    \begin{tabular}{c|c|cccccc}
    \Xhline{1.0pt}
    Module & Field & \#Params & {\tiny Complexity} & FLOPs & {\tiny Complexity} & MCA & MPCA\\
    \hline\hline
    \multirow{4}[2]{*}{\shortstack{DR+DW}} & 3×3 & 1.305M &\multirow{4}[2]{*}{\scalebox{.8}{\shortstack{$\Theta (D(3K^2\!$ \\ $+D))$}}} & 0.311G & \multirow{4}[2]{*}{\scalebox{.8}{\shortstack{$\Theta (T\!N\!D\!$ \\ $(3K^2\!+\!D))$}}} & \textbf{93.1} & \textbf{93.3} \\
          & 5×5  & 2.976M & & 0.712G & & 92.7 & 93.1  \\
          & 7×7  & 8.432M & & 2.021G & & 92.4 & 92.7  \\
          & 9×9  & 21.212M & & 5.089G & & 92.5 & 93.0  \\
    \hline
    \multirow{4}[2]{*}{\shortstack{ST factorised \\ DR+DW}} & 1×3, 3×1 & 2.160M &\multirow{4}[2]{*}{\scalebox{.8}{\shortstack{$\Theta(2\!D(3K\!$ \\ $+D))$}}} & 0.516G &\multirow{4}[2]{*}{\scalebox{.8}{\shortstack{$\Theta (2T\!N\!D\!$ \\ $(3K\!+\!D))$}}} & 92.6 & 93.0  \\
          & 1×5, 5×1 & 2.258M & & 0.540G & & 92.3 & 92.8  \\
          & 1×7, 7×1 & 2.406M & & 0.575G & & 92.8 & 93.0  \\
          & 1×9, 9×1 & 2.602M & & 0.622G & & 92.1 & 92.5  \\
    \hline
    \multirow{4}[2]{*}{\shortstack{Lite \\ DR+DW}} & 3×3 & 0.180M &\multirow{4}[2]{*}{\scalebox{.8}{\shortstack{$\Theta (D_l(3K^2$ \\ $+\!D_l\!+ D))$}}} & 0.042G &\multirow{4}[2]{*}{\scalebox{.8}{\shortstack{$\Theta (T\!N\!D_l\!$ \\ $(3K^2\!+\!D_l\!$ \\ $+D))$}}} & 92.6 & 92.8 \\
          & 5×5  & 0.387M & & 0.092G & & 92.6 & 93.1  \\
          & 7×7  & 1.069M & & 0.256G & & 92.3 & 92.7  \\
          & 9×9  & 2.667M & & 0.639G & & 92.3 & 92.5  \\
    \Xhline{1.0pt}
    \end{tabular}%
  \vspace{0.2cm}
  \caption{Results for \textbf{increasing initialized interaction fields} using three models. Backbone: ResNet-18. Computational cost for the backbone and embedding layer is \textbf{not} included. $K=9,25,49,81$ for 4 interaction fields and $D_l = \frac{D}{8}$.}
  \label{field_variance}%
  \vspace{-0.2cm}
\end{table}%

\textbf{Initialized Interaction Field for DIN and its Variants} \ To model the spatio-temporal interactions among people, an interaction field with appropriate size should be selected. We mainly provide with experiments on DIN and its two variants to choose an appropriate size: \textbf{i)} single interaction field that is initialized to cover a certain spatio-temporal domain, \textit{e.g.}, $3 \times 3$; \textbf{ii)} stacking layers that separately cover spatial and temporal domain (ST factorised model), \textit{e.g.}, $1 \times 3$ and $3 \times 1$; \textbf{iii)} lite model that covers a certain spatio-temporal domain. The results for increasing interaction field sizes are shown in Table \ref{field_variance}. It indicates that \textbf{i)} Larger interaction field sizes will not result in a good performance. \textbf{ii)} Similarly, stacking layers to separately model spatial and temporal interactions also result in slightly worse results. We consider they are due to the over-smoothing problem \cite{deeper_insignt_intoGCN} caused by stacking layers or too dense connections, which brings about excessive similarity between person features. \textbf{iii)} ST factorised and lite model both distinctly reduce the cost by reducing the exponent of $K$ and the value of $D$ to $D_l$, while both maintaining better results than previous methods.



\subsection{Comparisons with the State-of-the-Art}

\begin{table}[t]
  \small
  \centering
    \begin{tabular}{c|c|cc}
    \Xhline{1.0pt}
    Method & Backbone & MCA & MPCA \\
    \hline\hline
    SBGAR \cite{li2017sbgar} & Inception-v3 & 66.9  & 67.6  \\
    SSU \cite{bagautdinov2017social}   & Inception-v3 & 89.9  & - \\
    CERN-2 \cite{shu2017cern} & VGG-16 & 83.3  & 83.6  \\
    SPA+KD \cite{tang2018spa+kd} & VGG-16 & 89.3  & 89.0  \\
    PCTDM \cite{yan2018pctdm} & ResNet-18 & 90.3  & 90.5  \\
    stagNet \cite{qi2018stagnet} & VGG-16 & 89.3  & - \\
    CRM \cite{azar2019crm}   & I3D   & 92.1  & - \\
    ARG \cite{wu2019learning}  & ResNet-18 & 91.1  & 91.4  \\
    PRL \cite{hu2020prl}   & VGG-16 & 91.4  & 91.8  \\
    AT \cite{gavrilyuk2020actor} & ResNet-18 & 90.0  & 90.2  \\
    SACRF \cite{pramono2020eccvempowering} & ResNet-18 & 90.7  & 91.0  \\
    STBiP\textsuperscript{*} \cite{yuan2021learningcontext} & Inception-v3 & 91.3 & - \\
    HiGCIN \cite{yan2020higcin} & ResNet-18 & 91.4  & 92.0  \\
    \hline
    \multirow{2}[2]{*}{Ours-DIN} & VGG-16 & \textbf{93.6} & \textbf{93.8} \\
          & ResNet-18 & \textbf{93.1}  & \textbf{93.3}  \\
    \hline
    \multirow{2}[2]{*}{Ours-Lite DIN} & VGG-16 & \textbf{93.2} & \textbf{93.4} \\
          & ResNet-18 & \textbf{92.6}  & \textbf{92.8}  \\
    \Xhline{1.0pt}
    \end{tabular}%
  \vspace{0.2cm}
  \caption{Comparisons with previous state-of-the-art methods on \textbf{Volleyball dataset}. We mark with '-' if results are not provided. \textsuperscript{*} denotes results without visual context for fair comparison.} 
  \label{SOTA_Volleyball}%
\end{table}%

\begin{table}[t]
  \small
  \centering
    \begin{tabular}{c|cc}
    \Xhline{1.0pt}
    Method & Backbone & MPCA \\
    \hline
    \hline
    HDTM\cite{ibrahim2016hierarchical}  & AlexNet & 89.7  \\
    CERN-2\cite{shu2017cern} & VGG-16 & 88.3  \\
    Recurrent Modeling\cite{wang2017merge_two_class} & VGG-16 & 89.4 \\
    PCTDM\cite{yan2018pctdm} & AlexNet & 92.2  \\
    stagNet\cite{qi2018stagnet} & VGG-16 & 89.1  \\
    SPA+KD\cite{tang2018spa+kd} & VGG-16 & 92.5  \\
    ARG\cite{wu2019learning}   & ResNet-18 & 92.3  \\
    PRL\cite{hu2020prl}   & VGG-16 & 93.8  \\
    HiGCIN\cite{yan2020higcin} & ResNet-18 & 93.0  \\
    \hline
    \multirow{2}[1]{*}{Ours-DIN} & VGG-16 & \textbf{95.9} \\
          & ResNet-18 & \textbf{95.3}  \\
    \hline
    \multirow{2}[1]{*}{Ours-Lite DIN} & VGG-16 & \textbf{94.0} \\
          & ResNet-18 & \textbf{93.8}  \\
    \Xhline{1.0pt}
    \end{tabular}%
    \vspace{0.2cm}
    \caption{Comparisons with previous state-of-the-art methods on \textbf{Collective Activity datatset}.}
  \label{SOTA_Collective}%
\end{table}%

\begin{figure*}[t]
\centering
\includegraphics[width=1\textwidth]{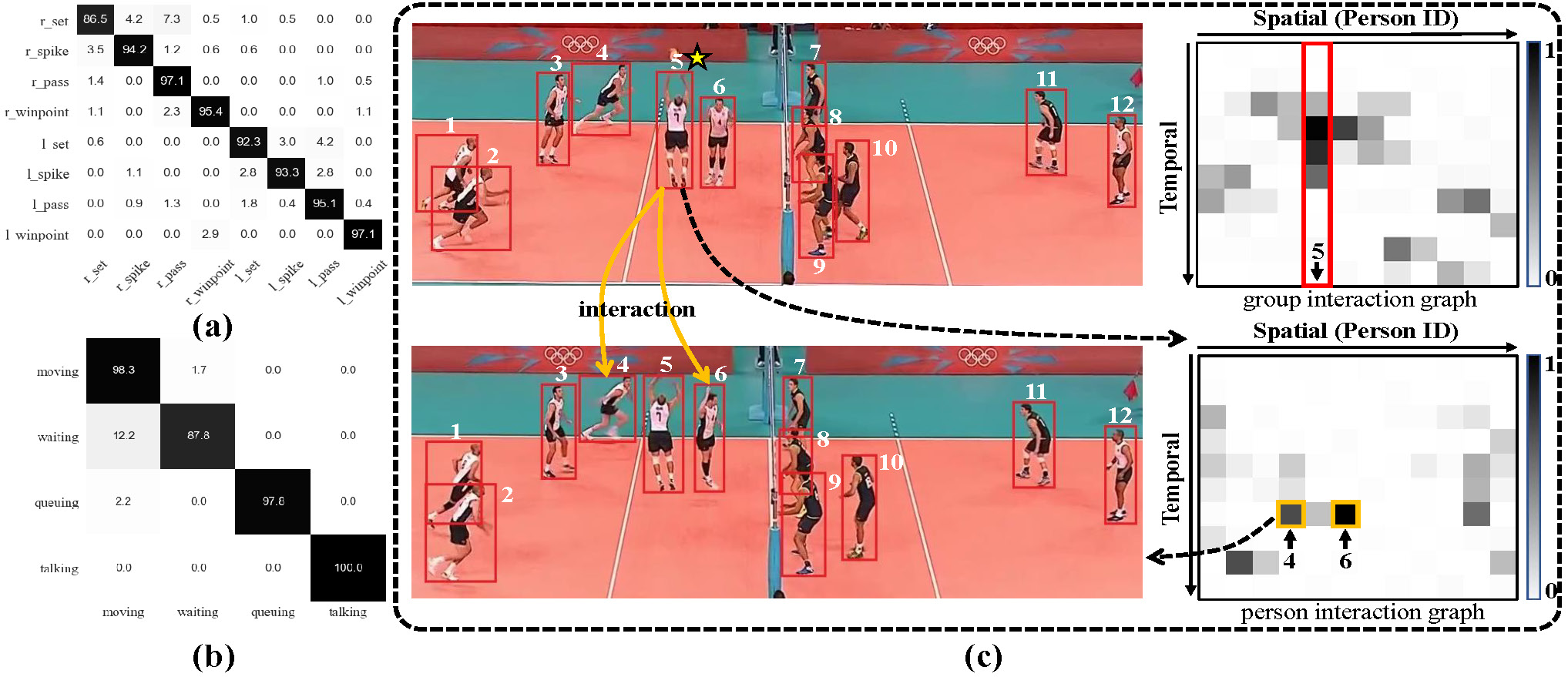} 
\caption{(a) \textbf{The confusion matrix for Volleyball dataset} using VGG-16 as a backbone. (b) \textbf{The confusion matrix for Collective Activity dataset} using VGG-16 as a backbone. (c) \textbf{Visualizations of a \textit{left set} activity example.} The upper left image is the starting image of the video clip. The upper right is the corresponding group interaction graph. The lower right is the interaction graph of the 5th person (key person, the red box in the group interaction graph). The lower left illustrates two of the 5th person's key interactions (yellow boxes in the 5th person's interaction graph).}
\label{STgraph_example}
\end{figure*}

In this subsection, we compare our models with previous state-of-the-art models. For a fair comparison, we only adopt RGB images as our model input and adopt a single backbone. 

\textbf{Performance on Volleyball dataset} \ The result is shown in Table \ref{SOTA_Volleyball}. Generally, our methods can achieve impressive results on this dataset. For methods using ResNet-18, our method can surpass them by 1.7\%. For methods using VGG-16, our methods can surpass them by 2.2\%. If considering the computational overhead of DR and DW, our models show more superiority. Our models generally cost less computational overhead and perform better than RNN-based models like \cite{yan2018pctdm,qi2018stagnet,bagautdinov2017social,li2017sbgar,shu2017cern}, due to a better representation from jointly modeling of spatial-temporal interaction. Our models outperform GNN-based methods like \cite{wu2019learning,yan2020higcin,gavrilyuk2020actor}, which ascribes to the dynamicity of the proposed modules. Method \cite{wu2019learning} even uses 16 graphs for reasoning but still trails our model. 

\textbf{Performance on Collective Activity dataset} \ The result is shown in Table \ref{SOTA_Collective}. With our proposed modules, the model with a VGG-16 backbone outperforms other methods by 2.1\% and a ResNet-18 backbone by 2.3\%. Note that Lite DIN which costs little extra computation can already achieve results on par with previous best methods, thus revealing the merits of introducing dynamicity.  

\textbf{Confusion Matrices} \ The confusion matrices of VGG-16 models on VD and CAD are respectively shown in Figure \ref{STgraph_example}(a) and (b). For VD, the modeling of dynamic spatial long-range interactions enables the model to distinguish left activities from right activities. Compared with confusion matrices from methods \cite{qi2018stagnet,yan2020higcin}, our methods performs well for \textit{pass} and \textit{set} activities. We ascribe it to the dynamic interaction modeling between spatio-temporal persons, because \textit{pass} and \textit{set} activities involve a person passing the ball and a person catching the ball. For CAD, compared with confusion matrices of methods \cite{ibrahim2016hierarchical,wang2017merge_two_class,yan2020higcin}, our methods distinguish the \textit{waiting} well. Previous methods mistake \textit{waiting} for \textit{moving} a lot because they fail to distinguish the temporal variations of people, which we tackle well.


\subsection{Qualitative Analysis}

\textbf{Group interaction graph} \ First, we visualize the group interaction graph for one example in the upper right image of Figure \ref{STgraph_example}(c), which sums all person-specific interaction graphs. It shows the people whom others interact more with to form the activity. If we sum along the temporal axis, we can find a key person (5th person, the red box in the group interaction graph) with the highest weight. In this example, it is the person performing \textit{setting} action, which is significant in the \textit{left set} group activity.  

\textbf{Person interaction graph} \ We take a step further by visualizing the key person's interaction graph in the lower right image of Figure \ref{STgraph_example}(c), which sums his interaction graphs in different temporal steps ($T$ graphs in total). It indicates that our modules enable global-level interactions though we initialize the interaction field locally. As shown in the person interaction graph, the yellow boxes are two of key interactions with the key person. In this example, they might spike the ball set from the key person.

\section{Conclusion and Future Works}
In this paper, we propose the Dynamic Inference Network to address the problems of inference on a predefined graph and inference in a computationally expensive way. With limited computation overhead, our model can achieve competitive results on publicly available datasets. Experiments have shown that person-specific interaction context is effective in inferring group activities. More challenging tasks and efficient inference models are left for future exploration. Moreover, this paper focus on the reasoning of person features, while a decent dynamic model incorporating visual context \cite{yuan2021learningcontext} are left for future exploration.

\noindent\textbf{Acknowledgement:} We would like to thank Jiayang Ren, Rong Jin and anonymous reviewers for their valuable feedback. This work was supported by the National Science Foundation China grant No. U1609213.

{\small
\bibliographystyle{ieee_fullname}
\bibliography{egbib}
}

\end{document}